\documentclass{article}

\usepackage[preprint]{neurips_2021}
\setcitestyle{numbers,square}
\usepackage[utf8]{inputenc} % allow utf-8 input
\usepackage[T1]{fontenc}    % use 8-bit T1 fonts
\usepackage{hyperref}       % hyperlinks
\usepackage{url}            % simple URL typesetting
\usepackage{booktabs}       % professional-quality tables
\usepackage{amsfonts}       % blackboard math symbols
\usepackage{nicefrac}       % compact symbols for 1/2, etc.
\usepackage{microtype}      % microtypography
\usepackage{xcolor}         % colors
\usepackage{amsmath}
\usepackage{subfigure}
\usepackage{graphicx}
\newcommand{\printfnsymbol}[1]{%
  \textsuperscript{\@fnsymbol{#1}}%
}

\title{Enhancing Hyperbolic Graph Embeddings via Contrastive Learning}

\newtheorem{definition}{Definition}
\author{%
  Jiahong Liu$^1$\thanks{Equal contribution; Work mainly done during an internship at Huawei Noah's Ark Lab.}, Menglin Yang$^2$\footnotemark[1], Min Zhou$^3$\thanks{Corresponding author.}, Shanshan Feng$^1$, Philippe Fournier-Viger$^4$
   
    \\
  $^1$Harbin Institute of Technology(Shenzhen); $^2$The Chinese University of Hong Kong; \\
  $^3$Huawei Technologies co.ltd; $^4$Shenzhen University. \\
  jiahong.liu21@gmail.com; mlyang@cse.cuhk.edu.hk; zhoumin27@huawei.com; \\ victor\_fengss@foxmail.com; philfv8@yahoo.com \\
}

\begin{document}

\maketitle
\vspace{-20pt}
\begin{abstract}
     Recently, hyperbolic space has risen as a promising alternative for semi-supervised graph representation learning. Many efforts have been made to design hyperbolic versions of neural network operations. However, the inspiring geometric properties of this unique geometry have not been fully explored yet.  The potency of graph models powered by the hyperbolic space is still largely underestimated.  Besides, the rich information carried by abundant unlabelled samples is also not well utilized. Inspired by the recently active and emerging self-supervised learning, in this study, we attempt to enhance the representation power of hyperbolic graph models by drawing upon the advantages of contrastive learning. More specifically, we put forward a novel Hyperbolic Graph Contrastive Learning (HGCL) framework which learns node representations through multiple hyperbolic spaces to implicitly capture the hierarchical structure shared between different views. Then, we design a hyperbolic position consistency (HPC) constraint based on hyperbolic distance and the homophily assumption to make contrastive learning fit into hyperbolic space. Experimental results on multiple real-world datasets demonstrate the superiority of the proposed HGCL as it consistently outperforms competing methods by considerable margins for the node classification task.
     \vspace{-10pt}
\end{abstract}

\section{Introduction}

Hyperbolic space, which is under the mathematical framework known as \textit{Riemannian geometry}~\cite{Riemannian}, has emerged as a promising alternative for graph representation learning recently~\cite{Chami2019hgcn,liu2019HGNN,zhang2019hgat, yang2021discrete,chen2021modeling}. Different from the Euclidean space which expands polynomially, the hyperbolic space grows exponentially
with its radius  which  can then be regarded as a smooth version of trees as the number of nodes in a binary tree also grows exponentially with the depth~\cite{bachmann2020constant}. Hence, it gains natural advantages in abstracting  scale-free graphs with a hierarchical organization~\cite{2010hyperbolic}. 

Inspired by the recently active and flourishing self-supervised learning, we aim at enhancing the representation power of hyperbolic graph models by drawing upon the advantages of contrastive learning. A contrastive learning algorithm commonly includes a loss function with a projection-powered (i.e., dot product) softmax function to maintain the consistency of the positives and negatives counterparts, which is found to be hardness-aware in nature and gives penalties to samples according to their hardness~\cite{uniformity}. Besides,  it is also found to be able to enforce extra intra-class more compactly and inter-class discrepancy simultaneously, leading to a better discriminative power of the models \cite{wang2021contrastive}.

Although the contrastive concept has been successfully utilized in Euclidean models~\cite{CG3,peng2020graph,zhu2020deep}, its application in hyperbolic space is hindered by the following two \textbf{challenges}. First, the embeddings optimized by traditional measure (inner product) in Euclidean space are with regularized norm, which makes it impossible to be pushed far away from the origin to utilize the spacious advantage of hyperbolic space. Second, the contrastive loss aims at preserving maximal information (i.e., \textit{uniformity}) during training~\cite{uniformity} by pushing all different instances apart and pulling positive pairs closer. This may destroy the prior structural relations of hierarchical datasets (i.e., \textit{tolerance})~\cite{tolerant}, which is detrimental to downstream tasks.

In this work, We introduce a hyperbolic graph contrastive learning framework (HGCL), which brings the benefits of contrastive learning into semi-supervised hyperbolic graph neural networks. Our contributions are summarized as follows:

\begin{itemize}
    \item A hyperbolic position consistency (HPC) constraint based on hyperbolic geometry is proposed to accommodate the two challenges mentioned above. It includes a positive  sampling  strategy balancing the tolerance and uniformity, and a distance-aware discriminator to properly measure the embeddings in hyperbolic space.

    \item Extensive experiments show that the proposed method outperforms the baseline models by large margins in the node classification task. The ablation study and analysis further gives insights into how the proposal successfully produces high-quality embeddings.
    
     \item To the best of knowledge, this is the first attempt to bridge contrastive learning with hyperbolic graph learning, shedding light to the related research topic.
\end{itemize}

\section{Preliminary}
\textbf{Hyperbolic Geometry.}
Hyperbolic geometry is a non-Euclidean geometry with a constant
    negative curvature. In this part, we briefly review the definitions  and concepts on hyperbolic geometry. A thorough and in-depth explanation can be found in \cite{lee2013smooth,nickel2018learning}.

%  A Riemannian manifold $(\mathcal{M},g)$ is a real and smooth manifold with a Riemannian metric $g_\mathbf{x}: \mathcal{T}_\mathbf{x}\mathcal{M}\times\mathcal{T}_\mathbf{x}\mathcal{M}\to \mathbb{R}$ at each point $\mathbf{x}\in \mathcal{M}$.
%     % 	a generalization of a 2-D surface with high dimensions, and can be locally approximated by $\mathbb{R}^n$. For each point $\mathbf{x}$ in $\mathcal{M}$, 
% 	The $\mathcal{T}_\mathbf{x}\mathcal{M}$ is the tangent space defined as the first-order approximation of $\mathcal{M}$ around $\mathbf{x}$, which is a $n$-dimensional vector space and isomorphic to $\mathbb{R}^n$.
%  For a point $\mathbf{x}\in \mathcal{M}$ and vector $\mathbf{v}\in\mathcal{T}_\mathbf{x}\mathcal{M}$, there exists a unique geodesic $\gamma:[0,1]\to\mathcal{M}$ where $\gamma(0)=\mathbf{x}, \gamma^\prime(0)=\mathbf{v}$. The exponential map $\exp_{\mathbf{x}}: \mathcal{T}_\mathbf{x}\mathcal{M} \to \mathcal{M}$ is defined as $\exp_{\mathbf{x}}(\mathbf{v})=\gamma(1)$ and logarithmic map $\log_\mathbf{x}$ is the inverse of $\exp_{\mathbf{x}}$. The parallel transport $PT_{\mathbf{x}\rightarrow \mathbf{y}}:\mathcal{T}_\mathbf{x}\mathcal{M}\to\mathcal{T}_\mathbf{y}\mathcal{M}$ achieves the transportation from point $\mathbf{x}$ to $\mathbf{y}$ that preserves the metric tensors. The geodesics is defined as $L(\gamma)=\int_\alpha^\beta\|\gamma^\prime(t)\|_g dt$, for $\gamma:[\alpha,\beta]\to \mathcal{M}$. Then the distance of $u, v \in \mathcal{M}$ is $d^\mathcal{M}(u,v)=\inf L(\gamma)$ where $\gamma$ is a curve such that $\gamma(a)=u, \gamma(b)=v$. 

    There are multiple hyperbolic models with different definitions and metrics that are mathematically equivalent. We here mainly consider two widely studied ones: Poincar\'e ball model \cite{nickel2017poincare} and the Lorentz model (also known as the hyperboloid model)~\cite{nickel2018learninglorentz}, which are defined by Definition~\ref{def:poincare} and~\ref{def:lorentz}, respectively. The related formulas and operations, e.g., distance, maps, and parallel transport are further summarized in Table~\ref{tab:operation} (Appendix.~\ref{apendix:operations}), where  $\oplus_{K}$ and $\operatorname{gyr}[.,.]$ 
   are the M\"obius addition~\cite{ungar2007hyperbolic} and gyration operator~\cite{ungar2007hyperbolic}, respectively.
   
\begin{definition}[Poincaré Ball Model]
The $n$-dimensional Poincaré ball model with negative curvature $K(K$<$0)$ is defined as a Riemannian manifold $(\mathbb{D}_{K}^{n},g_\mathbf{x}^{\mathbb{D}})$, where $\mathbb{D}_{K}^{n}=\left\{\mathbf{x}\in \mathbb{R}^{n}:\left\langle \mathbf{x}, \mathbf{x}\right \rangle_2 < -\frac{1}{K} \right \}$ is an open $n$-dimensional ball with radius $1/\sqrt{-K}$. Its metric tensor $g_\mathbf{x}^{\mathbb{D}_K}=(\lambda_\mathbf{x}^K)^2g^\mathbb{E}$, where $\lambda_\mathbf{x}^K=\frac{2}{1+K\|\mathbf{x}\|_2^2}$ is the conformal factor and $g^\mathbb{E}=I$ is the Euclidean metric.
\label{def:poincare}
\end{definition}
    
\begin{definition}[Lorentz Model]
 The $n$-dimensional Lorentz model (also named Hyperboloid model) with negative curvature $K(K<0)$ is defined as the Riemannian manifold $(\mathbb{H}_{K}^{n}, g_\mathbf{x}^{\mathbb{H}})$, where $\mathbb{H}_{K}^{n} = \left \{  \mathbf{x}\in \mathbb{R}^{n+1} : \left \langle \mathbf{x},\mathbf{x} \right \rangle_{\mathcal{L}}=\frac{1}{K} \right \}$ and $g_\mathbf{x}^{\mathbb{H}}=\mbox{diag}([-1,1,...,1])$. $\langle\cdot,\cdot\rangle$ is the Lorentzian inner product. 
 Let $\mathbf{x}, \mathbf{y} \in \mathbb{R}^{n+1}$, then the Lorentzian inner product is defined as:
\begin{equation}
\small
\begin{aligned}
    \left< \mathbf{x},\mathbf{y} \right>_\mathcal{L}:=-x_0y_0 + \sum_{i=1}^n x_iy_i.
\end{aligned}
\label{eq:hyperboloid}
\end{equation}

 \label{def:lorentz} 
\end{definition}

It is worth noting that the tangent space at $\mathbf{x}$ is given by a $n$-dimensional vector space
approximating $\mathbb{H}_K^n$, that is
\begin{equation}
\small
\begin{aligned}
    \mathcal{T}_\mathbf{\mathbf{x}}\mathbb{H}_K^n:=\{\mathbf{v}\in\mathcal{R}^{n+1}:\left< \mathbf{v},\mathbf{x} \right>_\mathcal{L}=0\}.
\end{aligned}
\label{eq:tangent}
\end{equation}

\section{Metohd}

\begin{figure*}[!h]
\centering
\includegraphics[width=0.999\textwidth]{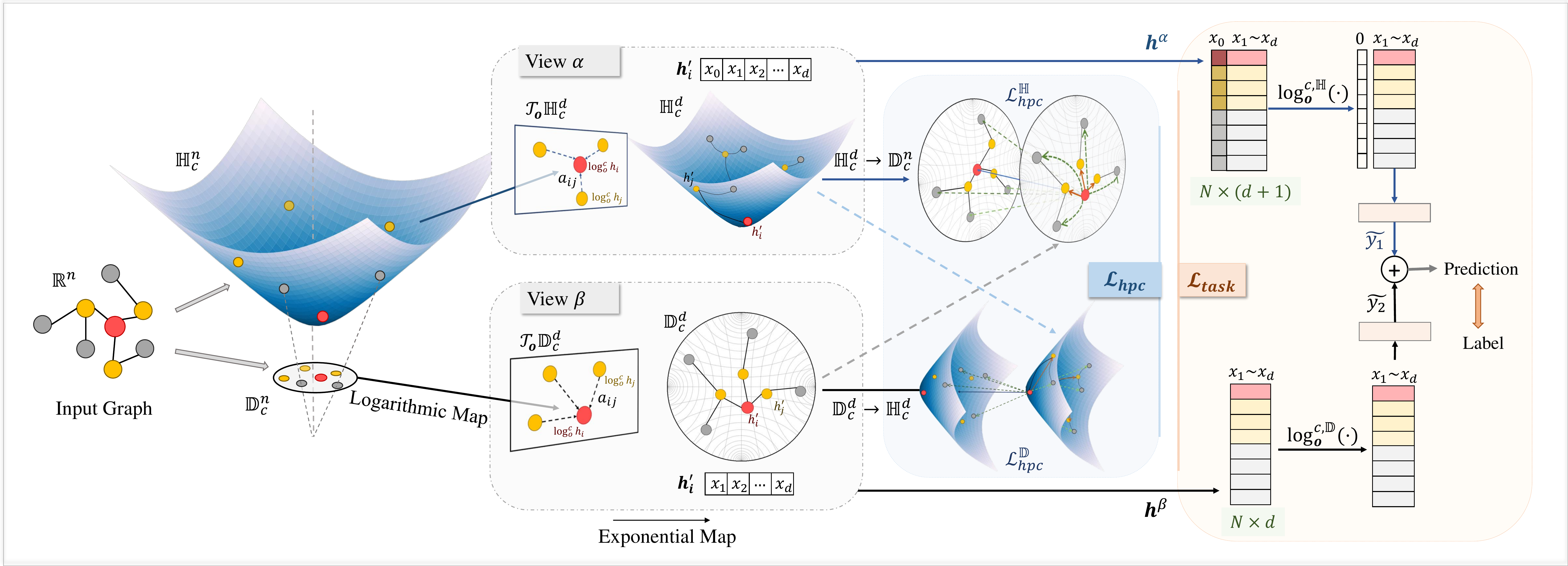}
\setlength{\abovecaptionskip}{-0.4cm}
\caption{Framework of HGCL. The input graph in Euclidean space is firstly mapped into two hyperbolic spaces. (i) Encoder: simultaneously propagate neighbor information in two HGNNs and obtain two node embeddings, i.e. $\mathbf{h}^{\alpha}$ and $\mathbf{h}^{\beta}$; (ii) Hyperbolic Contrastive Loss: refine the embeddings by pulling the positions of semantically similar samples closer and meanwhile pushing away the negative samples; (iii) Decoder and Task-specific Loss Function: decode the embeddings and obtain the final classification result during training.}
\label{fig:Framework}
\vspace{-10pt}
\end{figure*}

The proposed HGCL aims to enhance semi-supervised hyperbolic graph embeddings via contrastive learning, which is sketched in Figure~\ref{fig:Framework}. First, we encode the graph by two hyperbolic graph neural networks (HGNNs), producing different hyperbolic views, instead of tedious and tricky graph augmentations. Then, a hyperbolic contrastive loss, namely, hyperbolic position consistency (HPC) $L_{hpc}$ is further proposed to refine the embeddings, which is further decoded according to downstream tasks and trained with the task-specific loss function. We will mainly introduce the HPC, which is the key of our method.

\textbf{Hyperbolic Contrastive Loss: HPC.} In contrastive learning, the negative samples are required to avoid model collapse, and the same nodes in different views are usually selected as positive samples to extract discriminative features. Then how to select the samples and evaluate the pairs' similarities is critical. Available contrastive algorithms are customized for Euclidean space and are not directly applicable in hyperbolic space as hyperbolic space possesses distinctive properties (e.g, hierarchical awareness and spacious room) compared with the Euclidean counterpart. In this work, we propose a hyperbolic contrastive loss, i.e., HPC, which takes the one-hop neighbors of the anchors (center node) as positive samples regarding the homophily assumption to strengthen the tolerance, i.e., maintain semantic information. It further adds a penalty to negative samples with the purpose of pushing embeddings away from the origin to utilize the ample space near the boundary. More importantly, a distance-aware discriminator is designed to properly measure the similarities in hyperbolic space.

\vspace{-10pt}
\begin{figure}[htbp]
\centering
\includegraphics[width=0.53\textwidth]{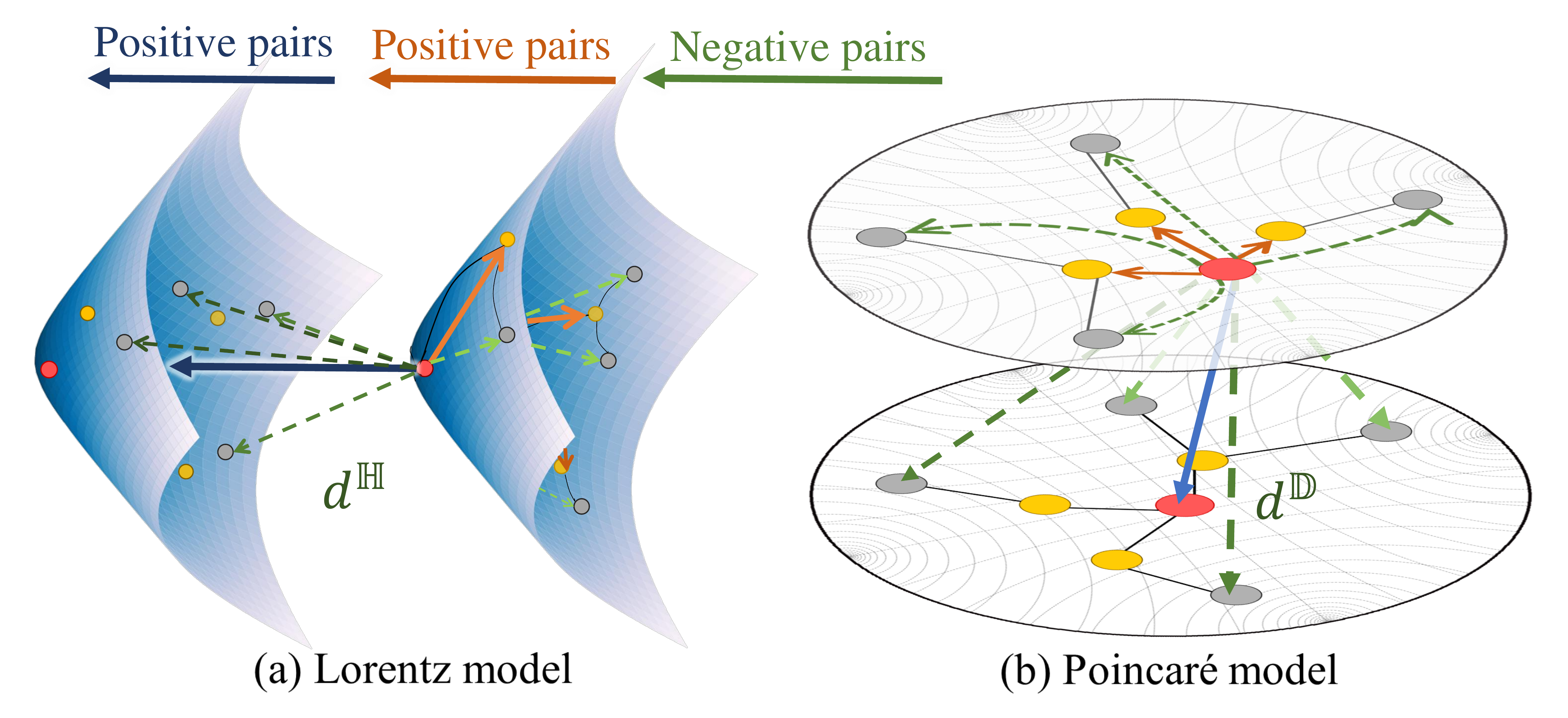}
\setlength{\abovecaptionskip}{-0.15cm}
\caption{Samples selection to calculate mutual information in different manifolds. The same nodes are in red and the neighbors in orange. The negative samples are also from two sources, which are intra-view and inter-view nodes (in green). }
\label{fig:MI}
\vspace{-7pt}
\end{figure}

\textbf{(1) Sample selection.} The main idea of sample selection is to enforce consistency between the encoded hyperbolic embeddings in different views and meanwhile maintain the semantic structure information. As illustrated in Figure~\ref{fig:MI}, for each anchor (center node), the positive samples consists of two black parts, which are the same node in the other view (named consistent node) and one-hop neighbors in the same view (named tolerance node).
To maximize the mutual information between similar nodes in principle, we define a local estimator $\text{MI}(\cdot,\cdot)$ based on the Jensen-Shannon divergence that distinguishes the \textit{positive} embeddings from \textit{negative} node embeddings.
Specifically, for each anchor $\mathbf{h}_i^\alpha$ in view $\alpha$, the two parts of pairwise loss are defined by Equation~\eqref{eq:JS_11} and Equation~\eqref{eq:JS_12}, respectively.

\begin{equation}
\small
\begin{split}
    \text{MI}\left(\mathbf{h}_i^\alpha, t_1(\mathbf{h}_i^\beta)\right)= \mathbb{E}_\mathbb{P}\left[\log \mathcal{D}\left(\mathbf{h}_i^\alpha, t_1(\mathbf{h}_i^\beta) \right) \right] + 
    \lambda_n \cdot \sum_{j=1}^m \mathbb{E}_{\mathbb{P}\times\bar{\mathbb{P}}}\left[\log \left(1-\mathcal{D} \left(\mathbf{h}_i^\alpha, \bar{t_1(\mathbf{h}_j^\beta)} \right) \right) \right],
\end{split}
\label{eq:JS_11}
\end{equation}

\begin{equation}
\small
\begin{split}
    \text{MI}\left(\mathbf{h}_i^\alpha, \mathcal{N}(\mathbf{h}_i^\alpha)\right)=\sum_{ \mathbf{h}_j \in \mathcal{N}(\mathbf{h}_i^\alpha)} \mathbb{E}_\mathbb{P}\left[\log \mathcal{D}\left( \mathbf{h}_i^\alpha, \mathbf{h}_j \right) \right] + 
    \lambda_n \cdot \sum_{j=1}^m \mathbb{E}_{\mathbb{P}\times\bar{\mathbb{P}}}\left[\log \mathcal{D}\left(1-\left(\mathbf{h}_i^\alpha, \bar{\mathbf{h}}_j^\alpha \right) \right) \right],
\end{split}
\label{eq:JS_12}
\end{equation}

where $\mathbf{h}_i^\alpha$ and $\mathbf{h}_i^\beta$ are embeddings of node $v_i$ in views $\alpha$ and $\beta$, respectively; $\mathcal{N}(\mathbf{h}_{_i}^\alpha)$ and $\mathcal{N}(\mathbf{h}_{_i}^\beta)$ are the embeddings of adjacent nodes $\mathbf{h}_i^\alpha$ and $\mathbf{h}_i^\beta$ in the same view; $t_1$ and $t_2$ are the functions that transfer the vectors from the manifold of view $\alpha$ to the manifold of view $\beta$; $m$ is the number of negative samples; $\lambda_n$ is a hyperparameter that adds a penalty to negative samples.

\textbf{(2) Distance-aware discriminator}
 As proxy for maximizing the local MI in Equation~\eqref{eq:JS_11} and \eqref{eq:JS_12}, we employ a discriminator $\mathcal{D}(\cdot,\cdot): \mathbb{R}^d \times \mathbb{R}^d \mapsto \mathbb{R}$, such that $\mathcal{D}(\mathbf{h}_i,\mathbf{h}_j)$ represents the probability scores assigned to the pair with two samples $\mathbf{h}_i$ and $\mathbf{h}_j$.  Note that, we implement the discriminator as the \textbf{distance} between two representations in corresponding space, $d_\mathbb{D}^K(\mathbf{h}_i, \mathbf{h}_j)$ or $d_\mathbb{H}^K(\mathbf{h}_i, \mathbf{h}_j)$ (see Table.\ref{tab:operation}) in Appendix.~\ref{apendix:operations}, instead of simply  using the dot product~\cite{2020Contrastive} or cosine similarity~\cite{2020Graph}.
 
 \begin{equation}
 \small
    L_{hpc} = - \frac{1}{2n} \sum_{i=1}^n \left[\underbrace{\text{MI}\left(\mathbf{h}_i^\alpha, t_1(\mathbf{h}_i^\beta) \right) + \text{MI}\left(\mathbf{h}_i^\beta, t_2(\mathbf{h}_i^\alpha) \right)}_{\text{$P_1$: Consistency}}
    + \underbrace{\text{MI}\left(\mathbf{h}_i^\alpha, \mathcal{N}(\mathbf{h}_i^\alpha) \right)
    + \text{MI}\left(\mathbf{h}_i^\beta, \mathcal{N}(\mathbf{h}_i^\beta) \right)}_{\text{$P_2$: Tolerance}}
    \right].
\label{eq:MI loss}
\end{equation}

\section{Experiments}
\textbf{Node classification.}
Following the standard practice and experimental setup in related work (HGCN)~\cite{Chami2019hgcn}, we report the F1-score for {\sc Disease} and {\sc Airport} datasets, and accuracy for the others in the node classification tasks. The statistics of datasets are listed in Appendix~\ref{subsection:statistics}. Table~\ref{tab:nc} shows the results, where the best records for each dataset have been marked in bold. Compared with baselines, the proposed HGCL achieves the best performance on all five datasets, indicating its powerful ability to embed graphs for node classification. It is thanks to the deployment of contrastive learning. Notably, our method not only performs well on datasets with lower hyperbolicity $\delta$ (e.g. {\sc Disease}, {\sc Airport}), but also shows significant improvements on those with higher $\delta$ (e.g. {\sc Cora}). Specifically, the accuracy is improved by 3.12\% on   {\sc Disease} and  4.05\% on {\sc Cora} compared with the second best, i.e.,   HGAT~\cite{zhang2019hgat}. What's more, the ablation study and further analysis are represented in Appendix~\ref{subsection:ablation} and Appendix~\ref{subsection:analysis}, respectively.
  
\begin{table}[htbp]  
\centering
\caption{Comparisons of node classification (NC) in accuracy with the standard deviation}
\resizebox{0.7\textwidth}{!}{%
\begin{tabular}{@{}lccccc@{}}
\toprule
Dataset   & {\sc Disease}    & {\sc Airport}    & {\sc PubMed}     & {\sc Citeseer}   & {\sc Cora}       \\ \midrule
EUC      & $32.56\pm1.19$  & $60.90\pm3.40$  & $48.20\pm0.76$  & $61.28\pm0.91$  & $23.80\pm0.80$  \\
HYP~\cite{nickel2017poincare}       & $45.52\pm3.09$  & $70.29\pm0.40$  & $68.51\pm0.37$  & $61.71\pm0.74$  & $22.13\pm0.97$  \\ \midrule
MLP       & $28.80\pm2.23$  & $68.90\pm0.46$  & $72.40\pm0.21$  & $59.53\pm0.90$  & $51.59\pm1.28$  \\

HNN~\cite{HNN}  & $41.18\pm1.85$  & $80.59\pm0.46$  & $69.88\pm0.43$  & $59.50\pm1.28$  & $54.76\pm0.61$  \\ %\midrule
%GCN       & $69.79\pm0.54$  & $81.59\pm0.61$  & $78.10\pm0.43$  & $70.35\pm0.41$  & $81.50\pm0.53$  \\
%GAT       & $70.40\pm0.49$  & $81.59\pm0.36$  & $78.21\pm0.44$  & $71.58\pm0.80$  & $83.03\pm0.50$  \\
%SAGE      & $70.10\pm0.49$  & $82.19\pm0.45$  & $77.45\pm2.38$  & $67.51\pm0.76$  & $77.90\pm2.50$  \\
%SGC       & $70.94\pm0.59$  & $80.59\pm0.16$  & $78.84\pm0.18$  & $71.44\pm0.75$  & $81.32\pm0.50$  \\ 

\midrule
HGNN~\cite{liu2019HGNN}     & $81.27\pm3.53$  & $84.71\pm0.98$  & $77.13\pm0.82$  & $69.99\pm1.00$  & $78.26\pm1.19$  \\
HGCN~\cite{Chami2019hgcn}       & $88.16\pm0.76$  & $89.26\pm1.27$  & $76.53\pm0.63$  & $68.04\pm0.59$  & $78.03\pm0.98$  \\
HGAT~\cite{zhang2019hgat}      & $90.30\pm0.62$  & $89.62\pm1.03$  & $77.42\pm0.66$  & $68.64\pm0.30$  & $78.32\pm1.39$  \\ \midrule
HGCL(ours) & $\textbf{93.42}\pm0.82$ & $\textbf{92.35}\pm1.01$ &   $\textbf{79.14}\pm0.68$   & $\textbf{72.11}\pm 0.64$ & $\textbf{82.37}\pm0.47$ \\ \bottomrule
\end{tabular}%
}
\vspace{-8pt}
\label{tab:nc}
\end{table}

\section{Conclusion}
In this work, we brought the benefits of contrastive learning into hyperbolic graph learning to obtain more powerful representations. In particular, we feed the graph-structured data into two hyperbolic encoders to generate contrastive views. Then, the proposed contrastive loss HPC leverages the properties of hyperbolic space to minimize the distances of positive-paired embeddings and maximize the distances of negative-paired embeddings. The extensive experimental results show that the contrast-powered learning scheme successfully preserves the semantic and hierarchies of the dataset as it consistently outperforms the baselines across the diverse datasets and tasks. As far as we know, this is the first hyperbolic graph learning framework powered by contrastive learning, which proposes a new direction for the research community.

\bibliographystyle{unsrt}  
\bibliography{references}  %%% Remove comment to use the external .bib file
\medskip
\newpage

\appendix

\section{Appendix}
\label{Appendix}
\subsection{Summary of Operations in Hyperbolic Models}
\label{apendix:operations}
\begin{table*}[htp!]
   % \small
    \centering
    \caption{Summary of operations in the Poincar{\'e} ball model and the Lorentz model ($K<0$)}
    \resizebox{1.00\textwidth}{!}{
    \begin{tabular}{ccc}
    \toprule
              & \textbf{Poincar{\'e} Ball Model} $(\mathbb{D}_K^n, g_\mathbf{x}^\mathbb{D})$ & \textbf{Lorentz Model $(\mathbb{H}_{K}^{n}, g_\mathbf{x}^{\mathbb{H}})$} \\ \midrule \midrule
\textbf{Distance}   &$d_{\mathbb{D}}^{K}(\mathbf{x}, \mathbf{y})=\frac{1}{\sqrt{|K|}} \cosh ^{-1}\left(1-\frac{2 K\|\mathbf{x}-\mathbf{y}\|_{2}^{2}}{\left(1+K\|\mathbf{x}\|_{2}^{2}\right)\left(1+K\|\mathbf{y}\|_{2}^{2}\right)}\right)$
    &  $d_{\mathbb{H}}^{K}(\mathbf{x}, \mathbf{y})=\frac{1}{\sqrt{|K|}} \cosh ^{-1}\left(K\langle \mathbf{x}, \mathbf{y}\rangle_{\mathcal{L}}\right)$  \\
    \textbf{Log map}    &$\log_{\mathbf{x}}^{K,\mathbb{D}}(\mathbf{y})=\frac{2}{\sqrt{|K|} \lambda_\mathbf{x}^{K}} \tanh ^{-1}\left(\sqrt{|K|}\left\|-\mathbf{x} \oplus_{K} \mathbf{y}\right\|_{2}\right) \frac{-\mathbf{x} \oplus_{K} \mathbf{y}}{\left\|-\mathbf{x} \oplus_{K} \mathbf{y}\right\|_{2}}$              
    &$\log_{\mathbf{x}}^{K,\mathbb{H}}(\mathbf{y})=\frac{\cosh ^{-1}\left(K\langle \mathbf{x}, \mathbf{y}\rangle_{\mathcal{L}}\right)}{\sinh \left(\cosh ^{-1}\left(K\langle \mathbf{x}, \mathbf{y}\rangle_{\mathcal{L}}\right)\right)}\left(\mathbf{y}-K\langle \mathbf{x}, \mathbf{y}\rangle_{\mathcal{L}} \mathbf{x}\right)$\\ 
    \textbf{Exp map}   & $\exp_{\mathbf{x}}^{K, \mathbb{D}}(\mathbf{v})=\mathbf{x} \oplus_{K}\left(\tanh \left(\sqrt{|K|} \frac{\lambda_{\mathbf{x}}^{K}\|\mathbf{v}\|_{2}}{2}\right) \frac{\mathbf{v}}{\sqrt{|K|}\|\mathbf{v}\|_{2}}\right)$
    & $\exp_{\mathbf{x}}^{K,\mathbb{H}}(v)=\cosh \left(\sqrt{|K|}\|\mathbf{v}\|_{\mathcal{L}}\right) \mathbf{x}+\mathbf{v} \frac{\sinh \left(\sqrt{|K|}\|\mathbf{v}\|_{\mathcal{L}}\right)}{\sqrt{|K|}|| \mathbf{v}||_{\mathcal{L}}}$ \\
    \textbf{Transport} &  $P T_{\mathbf{x} \rightarrow \mathbf{y}}^{K,\mathbb{D}}(\mathbf{v})=\frac{\lambda_{\mathbf{x}}^{K}}{\lambda_\mathbf{y}^{K}} \operatorname{gyr}[\mathbf{y},-\mathbf{x}] \mathbf{v} $& $ P T_{\mathbf{x} \rightarrow \mathbf{y}}^{K,\mathbb{H}}(\mathbf{v})=\mathbf{v}-\frac{K\langle \mathbf{y}, \mathbf{v}\rangle_{\mathcal{L}}}{1+K\langle \mathbf{x}, \mathbf{y}\rangle_{\mathcal{L}}}(\mathbf{x} + \mathbf{y})$ \\
    \bottomrule
    \end{tabular} 
    }
    \label{tab:operation}
    \end{table*}

\subsection{Ablation Study}
\label{subsection:ablation}
We conduct an ablation study to verify the effectiveness of HPC and its main components, i.e. positive sampling strategy and distance-aware discriminator. In particular, positive sampling strategy is simply removed (denoted as w/o pos), while the pairs' similarity is measured by the inner product (denoted as  w/o dis). %Besides, to shed light on contributions of the main components,i.e. positive sampling strategy and distance-aware discriminator,  .  
The results are summarized in Table ~\ref{tab:ablation study}. 

\begin{table}[!h]
\caption{Ablation study of the HPC on NC task.}
\label{tab:ablation study}
\centering
\resizebox{0.85\textwidth}{!}{%
\begin{tabular}{lccccc}
\toprule
        Method
        & {\sc Disease} & {\sc Airport} & {\sc Pubmed} &  {\sc Citeseer} &  {\sc Cora} \\ \midrule
w/o HPC & $90.79\pm1.83$                 & $91.98\pm1.15$                 & $75.77\pm0.47$                & $68.89\pm1.24$                  & $79.32\pm0.65$              \\
w/o pos & $91.84\pm1.04$                 & $92.12\pm1.06$                 & $75.83\pm0.67$                & $68.11\pm1.74$                  & $79.18\pm1.03$              \\
w/o dis & $90.76\pm2.19$                 & $92.00\pm0.95$                 & $75.46\pm0.78$                & $68.68\pm0.80$                  & $78.90\pm1.21$              \\ \midrule
ours    & $\mathbf{93.42}\pm0.82$                 & $\mathbf{92.35}\pm1.01$                 & $\mathbf{79.14}\pm0.68$                & $\mathbf{72.11}\pm0.64$                  & $\mathbf{82.37}\pm0.47$              \\ \bottomrule
\end{tabular}
}
\end{table}

As observed, the performance decreases significantly if HPC is removed, which confirms the function of HPC in our framework.  The results also reveal that both two components of HPC make essential endowment to boost the performance as any one is removed or changed from HPC is counterproductive in many cases. It is quite straightforward as dot product misjudges the similarity of hyperbolic embeddings and also hinders the utilization of the hyperbolic space.

\subsection{Visualization and Analysis}
\label{subsection:analysis}
% Figure~\ref{fig:hdo} further shows the distribution of the hyperbolic distance to origin (HDO) of embeddings in Poincar\'e ball of the {\sc airport} dataset for the link prediction task. %We analyze the changes of HDO between HGCN and HGCL, 
% As it observed, nodes of HGCL are embedded far away from the origin than the HGCN, which indicates that our proposal is able to push the embeddings to the boundary, taking advantage of the exponentially increasing space of hyperbolic models.

\begin{figure}[h]
\centering
\includegraphics[width=0.46\textwidth]{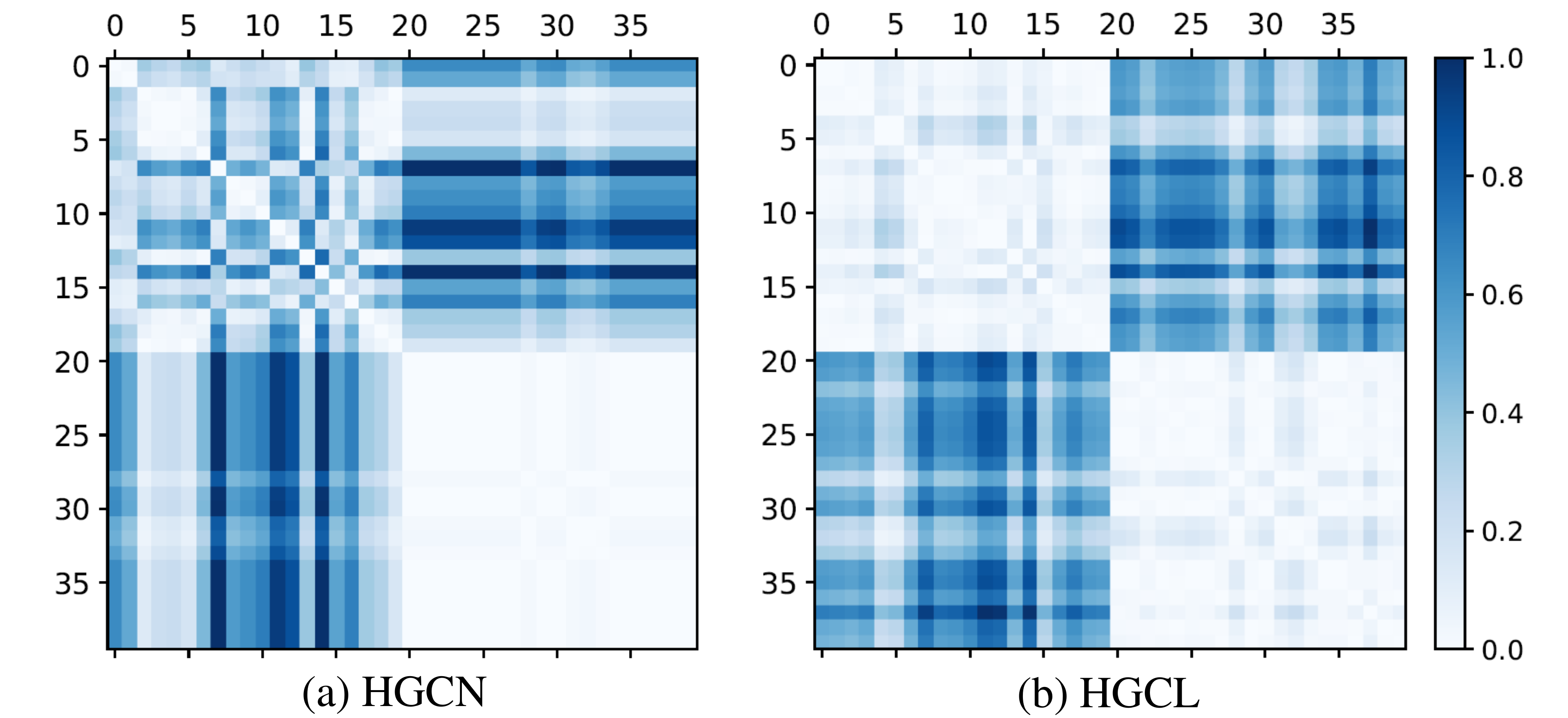}
\caption{The distance heatmap among inter-class and intra-class embeddings on {\sc Disease}}
\label{fig:class}
\end{figure}

Figure~\ref{fig:class} shows the distance heatmap of inter-class and intra-class embeddings on the {\sc Disease} dataset, where nodes 0-19 and 20-39 belong to  two different classes.
Compared to HGCN, it is observed that the node embeddings of HGCL in the same class are more compact, and the boundaries of different classes are less ambiguous. The observation is more evident for nodes 0-19 as the inter-class distances of the embeddings obtained by HGCN are much larger than those produced by HGCL. Thus, the contrast-enhanced hyperbolic leaning framework could pull representations of similar nodes together while pushing away the dissimilar ones, which confirms our intention that the well-designed contrastive loss is able to improve the discriminative power of hyperbolic models.

% \begin{figure}[h]
% \centering
% \includegraphics[width=0.5\textwidth]{HDO_0.pdf}
% \caption{Distribution of the hyperbolic distance to the origin (HDO)  embedded into the Poincar\'e ball on {\sc airport} dataset}
% \label{fig:hdo}
% \end{figure}

\subsection{Datasets}
\label{subsection:statistics}
Three types of networks (i.e., citation network, disease spreading network, and flight network) are used. The citation networks, including {\sc Cora}, {\sc Citeseer}, and {\sc PubMed}, are standard benchmark datasets widely used to evaluate the performance of graph-related models. Table.~\ref{tab:statistics} gives the statistics of the datasets.

\begin{table}[htbp]
\centering
\caption{Statistics of the datasets.}% used in the experiments.}
\resizebox{0.55\textwidth}{!}{%
\begin{tabular}{@{}lccccc@{}}
\toprule
{\sc Dataset}  & Nodes & Edges & Classes & Feature &Hyperbolicity $\delta$ \\ \midrule
{\sc Disease}  & 1044  & 1043  & 2       & 1000  &   0      \\
{\sc Airport}  & 3188  & 18631 & 4       & 4  &   1        \\
{\sc Citeseer} & 3327  & 4732  & 6       & 3703  &   3.5      \\
{\sc PubMed}   & 19717 & 88651 & 3       & 500   &  3.5      \\
{\sc Cora}     & 2708  & 5429  & 7       & 1433  & 11       \\ \bottomrule
\end{tabular} 
}
\label{tab:statistics}
\end{table}

\end{document}